\documentclass[a4paper, 10pt, onecolumn]{article}

\usepackage{etoolbox}	
\newtoggle{NAT}
\toggletrue{NAT}	

\usepackage{titling}
\usepackage[style=nature, 
	backend=bibtex,
	autocite=superscript, 
	citestyle=nature,
	doi = \iftoggle{NAT}{false}{true},
	sortcites=true,
	backref=true,
	date = year]{biblatex}
\renewcommand{\textcite}[1]{\citeauthor{#1}\,\supercite{#1}}
\usepackage[utf8]{inputenc}	

\usepackage{lmodern}
\usepackage[T1]{fontenc}
\usepackage[babel=false]{microtype}

\usepackage{tabularx}
\newcommand\setrow[1]{\gdef\rowmac{#1}#1\ignorespaces}
\newcommand\clearrow{\global\let\rowmac\relax}
\clearrow

\usepackage{amsmath}	
\usepackage{amssymb}	
\usepackage[affil-it]{authblk}	
\usepackage[title]{appendix}	
\usepackage{bm}			
\usepackage{booktabs}	
\usepackage[font={small}, labelfont={bf}]{caption}	

\let\counterwithin\relax
\usepackage{chngcntr}	
\usepackage{color}	
\usepackage{amsfonts} 
\usepackage{csquotes}	
\usepackage{paralist}	
\usepackage{graphicx}	
\usepackage{mathrsfs}	
\usepackage[group-separator={\,}]{siunitx}	
\usepackage{stfloats}	
\usepackage{subfig}		
\usepackage{tabularx}	
\usepackage[normalem]{ulem} 		
\usepackage{units}		

\usepackage[default]{gfsneohellenic}
\usepackage[LGR,T1]{fontenc} 

\usepackage[hyperfootnotes=false]{hyperref}		
\usepackage[all]{hypcap}
\usepackage{xcolor}
\definecolor{dark-red}{rgb}{0.45,0.15,0.15}
\definecolor{dark-blue}{rgb}{0.15,0.15,0.4}
\definecolor{medium-blue}{rgb}{0,0,0.5}
\hypersetup{
    colorlinks, 				
    linkcolor={dark-red},		
    citecolor={dark-blue},		
    urlcolor={medium-blue},		
}

\newcommand*{\bv}[1]{\mathbf{#1}}	

\newcommand{\K}{K}

\newcommand{\xVec}{\mathbf{x}}

\newcommand{\Xt}{{\bv X}}

\newcommand{\Rt}{{\bv R}}
\newcommand{\rn}{{\bv r}_n}

\newcommand{\rnoise}{{ \hat{ \bv r}}_n}

\renewcommand{\r}{{{\bv r}}}

\newcommand{\pare}[1]{ \left(#1\right)}

\newcolumntype{Y}{>{\raggedright\arraybackslash}X}

\graphicspath{{.}{./figures-to-upload/}}

\definecolor{burgundy}{rgb}{0.5, 0.0, 0.13}

\title{\textbf{Learning physical properties of anomalous random walks using graph neural networks
}}

\newcommand{\affilPasteur}{Decision and Bayesian Computation, USR 3756 (C3BI/DBC) $\&$ Neuroscience department CNRS UMR 3751, Institut Pasteur, CNRS, Paris, France}

\newcommand{\affilSanofi}{Histopathology and Bio-Imaging Group, Sanofi R\&D, Vitry-Sur-Seine, France}

\newcommand{\correspondingauthors}{Correspondence should be addressed to hverdier@pasteur.fr $\&$ jbmasson@pasteur.fr}

\newcommand{\affilUDP}{Universit\'e de Paris, UFR de physique, 75013 Paris, France}

\author[1,2,3,*]{Hippolyte Verdier}
\author[1]{Maxime Duval}
\author[1]{Fran\c{c}ois laurent}
\author[2]{Alhassan Cass\'e}
\author[1]{Christian L. Vestergaard}
\author[1,*]{Jean-Baptiste Masson}
\affil[*]{\correspondingauthors}
\affil[1]{\affilPasteur}
\affil[2]{\affilSanofi}
\affil[3]{\affilUDP}

\begin{document}

\maketitle

\begin{abstract}

Single particle tracking allows probing how biomolecules interact physically with their natural environments. 
A fundamental challenge when analysing recorded single particle trajectories is the inverse problem of inferring the physical model or class of models of the underlying random walks. 
Reliable inference is made difficult by the inherent stochastic nature of single particle motion, by experimental noise, and by the short duration of most experimental trajectories.
Model identification is further complicated by the fact that main physical properties of random walk models are only defined asymptotically, and are thus degenerate for short trajectories. 

Here, we introduce a new, fast approach to inferring random walk properties based on graph neural networks (GNNs). Our approach consists in associating a vector of features with each observed position, and a sparse graph structure with each observed trajectory. By performing simulation-based supervised learning on this construct~\cite{Cranmer2020}, we show that we can reliably learn models of random walks and their anomalous exponents. 
The method can naturally be applied to trajectories of any length. 
We show its efficiency in analysing various anomalous random walks of biological relevance that were proposed in the AnDi challenge~\cite{MuozGil2020}. 
We explore how information is encoded in the GNN, and we show that it learns relevant physical features of the random walks.
We furthermore evaluate its ability to generalize to types of trajectories not seen during training, 
and we show that the GNN retains high accuracy even with few parameters.
We finally discuss the possibility to leverage these networks to analyse experimental data.

 
\end{abstract}

\section{Introduction}

Random walks are encountered throughout a large variety of scientific domains, spanning diverse fields such as condensed matter physics~\cite{hughes1995random}, molecular biology~\cite{Wachsmuth2000}, ecology~\cite{Sims2008}, and finance~\cite{shreve2004stochastic}. The random walkers can be considered probes of their environment, and their recorded trajectories contain information on the properties of both the walkers and their environments. 
When analysing experimental data from such systems, one needs to solve the inverse problem of inferring an appropriate model or class of models and their parameters from a given set of trajectories. 

Whether the recorded trajectories come from single-molecule experiments~\cite{Betzig2006,Sieben2018} or from other types of experimental recordings~\cite{Maiuri2015,Cavagna2010,OToole2015}, the inverse problem of identifying the nature (i.e.\ model class and parameters) of the random walks is a complex one. 
The reasons for this complexity are multiple:
the inherent stochastic nature of random walks makes it difficult to identify stereotypical behaviours, 
characteristic features of different random walks models overlap for finite size trajectories, 
experimental noise can mask subtle local dynamics that could differentiate some models, 
and the distributions of selective features associated with many random walks are difficult to compute.

A continually growing number of approaches have been developed to tackle the task of identifying and characterising random walks from their individual realisations. 
A popular class of approaches consists in fitting temporal curves of moments of the displacements~\cite{Qian1991, Weber2012}. 
The main such statistics is the mean square displacement (MSD), either estimated from an ensemble of random walks or evaluated for individual trajectories~\cite{Young2017,Kepten2015,Lanoisele2018,Qian1991}. The MSD is widely used and remains a tool of choice for analysis, both for historical reasons and for its simple interpretation~\cite{hughes1995random} (see Section~\ref{rationale} below). 
The MSD however has several undesirable statistical properties which can make conventional analysis suboptimal. It is nonstationary, highly autocorrelated, and has a non-Gaussian distribution~\cite{Vestergaard2014}, making the design of efficient inference procedures difficult. Moreover, it may be insufficient on its own to distinguish between different random walk models, especially for heterogeneous systems~\cite{Kepten2013, Metzler2014}. 
The inferences of the model class and associated parameters degrade with decreasing trajectory lengths, with increasing positioning noise and with positional correlations induced by the experimental measurement process.
Many complementary approaches, based on predefined features extracted from trajectories, have been developed. Some are ensemble-based, relying on distributions of features evaluated in sets of trajectories, such as the distribution of tangent velocities~\cite{Burov2013}, of displacements~\cite{Schtz1997}, of "excursions"~\cite{Tejedor2010}, or of first time passage statistics~\cite{Condamin2008}. 
Others are centred on individual trajectories, such as non-ergodicity estimators~\cite{Lanoisele2016}, renormalisation-group-based  second moment analysis~\cite{OMalley2012}, mixed use of multiple estimators in decision trees~\cite{Meroz2015}, asymmetry samplers~\cite{Saxton1993}, or filtered temporally-averaged MSDs~\cite{Burnecki2015}. 

Statistical learning has provided new tools 
to address these inverse problems in a principled way. 
Hidden Markov Models (HMM)~\cite{christopherbishop2011} are efficiently used to probe diffusion modes~\cite{Monnier2015,Das2009,Lindn2018}, detect confined-like motion~\cite{Slator2018}, and mix local and global information to perform inference on very short trajectories~\cite{Sgouralis2017}. 
(It is worth noting here that the temporally averaged MSD (TAMSD) can also be used in detecting diffusion mode changes~\cite{Grebenkov2019}.)
Other Bayesian approaches have allowed to infer fBM characteristics \cite{Hinsen2016,Krog2018,LochOlszewska2020} and to perform model selection between several different models~\cite{Koo2016}.  

The main drawback of statistical learning approaches is that they often require an analytical expression for the likelihood function.
This limits the applicability of these approaches as many random walk models of interest do not have a tractable likelihood. 

Recently, machine learning approaches have been developed to tackle this problem and to provide a broader class of random walks with efficient means for model selection and inference. 
To overcome the lack of ground truth annotations for experimental random walks, applications of machine learning generally involve a simulation-based approach, where a machine learning model is trained on simulated data from a selection of random walk models using supervised learning. 
Assuming that the selected training set contains models that describe the experimental system well enough,
the learned model can then 
be transferred to classify and/or infer the parameters of experimental random walks.
This differs from the typical approach to supervised learning where the model is trained on an annotated experimental dataset.
Within this type of approaches, ensemble learning methods such as random forest have been found to be remarkably effective when applied  
to appropriately selected features of recorded trajectories~\cite{1902.07942,Wagner2017,MuozGil2020}. 
Neural networks applied on windowed MSDs have also been found to be efficient for differentiating between modes of motion and detecting their transitions in time~\cite{Dosset2016}.

As larger datasets become available, deep learning approaches, which automatically learn features from the raw data, have consistently been found to outperform classic machine learning models relying on predefined features~\cite{LeCun2015}. 
Following this trend, deep learning approaches have recently been developed to infer properties of random walks. 
Recurrent neural networks (RNNs), tools of choice for time series analysis, have been used to infer parameters associated to sub-diffusive behaviour~\cite{lstm} and to detect changes between different diffusion modes~\cite{Arts2019}. 
An approach based on a deep learning architecture search platform, leveraging convolutional neural networks (CNNs), LSTM networks (a class of RNNs), and the ResNet and InceptionTime
architectures known for their efficiency for natural language processing (NLP)~\cite{vanKuppevelt2020}, 
has proven to be efficient even for short trajectories~\cite{1902.07942}.
Another recent approach, 
inspired from generative audio generation models~\cite{WaveNet}, leverages a temporal convolution network~\cite{TCN-vs-LSTM} where convolutions are applied with various dilation effect to detect long term correlations, has shown impressive performance in inferring the nature of random walks~\cite{wavnet-spt}.

The limitations of deep learning approaches applied to random walks are similar to other applications of deep learning:
identifying overfitting can be difficult, and learned models typically lack interpretability (they are usually referred to as black box models for this reason).
Furthermore, the computational cost of training generally grows with the number of parameters and with the level of recurrence of the neural network. 

The particular nature of random walk data and the simulation-based approach also adds difficulties when adapting to experimental data:
varying trajectory lengths complicate the application of non-recurrent neural networks (e.g., CNN architectures), and
varying experimental noise as well as mislinking between particle positions in experimental images
may degrade statistical performance. 
Finally, inevitable differences between our random walk models and the true data generating process (known as distribution drift, but here exacerbated by the simulation-based approach), 
makes it hard to define the conditions under which transfer learning performs well.  

To address some of these problems, we here propose to leverage graph neural networks (GNNs)~\cite{1810.00826}, a recent extension of neural networks to general dependency structures between data points (Fig.~\ref{fig:connection}).
Our goal is two-fold:

\begin{figure}[!h]
 \centering
  \includegraphics[width=.95\linewidth]{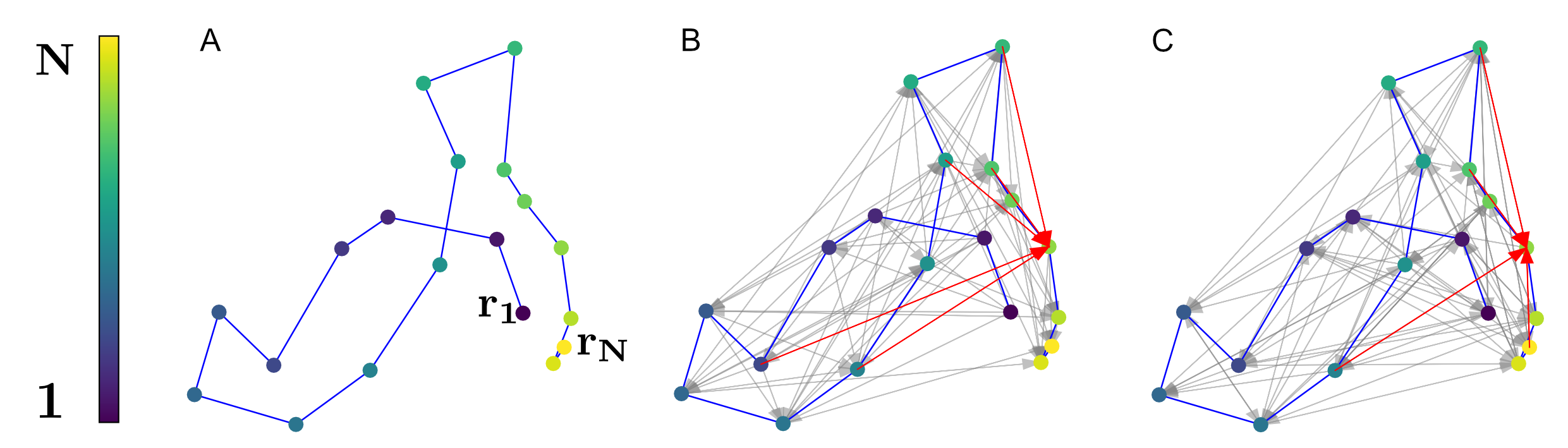}
  \caption{Examples of graphs associated to a single trajectory.
  A) An example of a short, subdiffusive fBM trajectory. 
  B) Graph associated to the trajectory following a causal geometric edge wiring scheme (see Section~\ref{graph-traj}). 
  C) Graph obtained by wiring edges at random.
  Edges linking to a single selected node are shown in red in B and C.}
  \label{fig:connection}
\end{figure}

(i)~ To develop an inference procedure that can be applied to trajectories of any length, is robust w.r.t.\ noise, and is numerically efficient, while keeping the number of parameters of the neural network low.  

(ii)~ To provide physical interpretability of the neural network model through analysis of the latent representations it learns, and to study its robustness w.r.t.\ differences in the statistical properties of the training and testing data which is essential for  transfer to experimental data analysis.

In this paper, we will focus our approach and applications on a limited subset of random walk models. 
However, the approach is transferable to any type of random walk.
We focus here on anomalous motion stemming from the following 5 different models, which were selected for the anomalous diffusion (AnDi) challenge~\cite{andi2020}: 
subdiffusive continuous time random walks (CTRW)~\cite{Scher1975,Magdziarz2009}, 
fractional Brownian motion (fBM)~\cite{Mandelbrot1968}, 
superdiffusive Levy walks (LW)~\cite{Klafter1994,Palyulin2014,Koren2007},
annealed transient time motion (ATTM)~\cite{Massignan2014}, 
and scaled Brownian motion (sBM)~\cite{Lim2002,Jeon2014,Sposini_2019}. 
We will also consider Brownian motion and the Ornstein-Uhlenbeck process~\cite{Gardiner} in the analysis of the representations learned by the  GNN.

Although the GNNs will always be trained on random walks whose nature does not change with time (i.e.\ each trajectory is generated by a single model with constant parameter values), we will explore the robustness of these trained networks to changes in the underlying generative model over time. 
We refer the reader to Refs.~\cite{ElBeheiry2015, Frishman2020, Serov2020, Hoze2017, Laurent2019} for methods addressing spatial and temporal heterogeneities in the overdamped Langevin equation and to Refs.~\cite{1902.07942, LochOlszewska2020} for fBM.

The paper is organised as follows. In Section~\ref{rationale} we briefly introduce  the different models of random walks and the motivation for our learning procedure.  In Section~\ref{graphs} we introduce the graph neural network approach to random walk inference, the graph building process and properties of the architecture. In Section~\ref{results} we discuss the results, quantify the properties of the learning procedure and explore what has been learnt by the network. We also explore neural network design choices and their effect on learning. Finally, in Section~\ref{discussion} we discuss the efficiency of the approach, challenges in its application to experimental data and possible extensions to unsupervised learning.

\section{Anomalous random walks}
\label{rationale}

Here we briefly introduce important properties of the anomalous random walk models~\cite{hughes1995random,Gardiner} that we will consider in the following. 
Anomalous diffusion emerges when a random walker diffuses in a disordered system or in an out-of-equilibrium environment. 
Usually, the principal property used to characterise an anomalous diffusion process is the temporal scaling behaviour of its mean squared displacement, 
\begin{equation}
    \langle \bf{r}^{2}\left(t\right) \rangle \propto t^{\alpha} \enspace,
\end{equation}
where $\bf{r}$ is the position of the random walker, $\langle \cdot \rangle$ denotes ensemble averaging, and $\alpha \in \left[0, 2\right]$ is the anomalous exponent. 
The process is subdiffusive when $\alpha < 1$, superdiffusive when $\alpha>1$ and Brownian (i.e.\ non anomalous) when $\alpha=1$. 

Anomalous diffusion is associated to at least one of the following properties:
(i) random walkers or their environment exhibit spatially or temporally varying properties, 
(ii) displacements are not statistically independent at any sufficiently small time scale, 
and 
(iii) displacements at small time scales exhibit anomalies that prevent the central limit theorem 
applying~\cite{Metzler2014}.

We consider the five following models of anomalous diffusion:

\begin{enumerate}
    \item Fractional Brownian motion (fBM)~\cite{Gardiner,Mandelbrot1968}. 
fBM is a Gaussian process characterised by long temporal correlations in the noise driving the process. 
It is generated by a Langevin equation~\cite{Gardiner} of the form $\frac{d\bf{r}\pare{t}}{dt} = \sqrt{\K_{\alpha}} \pmb{\eta}\pare{t}$, where $\pmb{\eta}$ is a zero-mean Gaussian noise process with covariance structure $\langle\pmb{\eta}\pare{t_{1}}\pmb{\eta}\pare{t_{2}}\rangle= \alpha\pare{\alpha-1}|t_{1}-t_{2}|^{\alpha - 2}$, and $\K_{\alpha}$ is a generalized diffusion constant that sets the scale of the process.
fBM is a self-similar Gaussian process with stationary increments~\cite{Gardiner,Mandelbrot1968}. Its associated noise is anti-persistent and negatively correlated in the subdiffusion regime but persistent and positively correlated  in the superdiffusion regime. 
As such, this random walk model displays anomalous property (ii) as defined above.
It is stationary and ergodic~\cite{Deng2009}, and its likelihood is analytically tractable.

\item Scaled Brownian motion (sBM)~\cite{Saxton2001,Lim2002}.
sBM is generated by a Langevin equation~\cite{Gardiner} 
with a  time-dependent diffusion coefficient~\cite{Saxton2001,Lim2002} of the form $K\pare{t}=\alpha K_{\alpha} t^{\alpha-1}$ and driven by an uncorrelated (white) noise. 
It can generate both subdiffusive and superdiffusive processes. 
sBM displays anomalous property (i), 
it is weakly non-ergodic~\cite{Sposini_2019} and has the same marginal probability density for the time-evolution of the walker’s position as the fBM~\cite{Jeon2014}, but a different autocorrelation structure.

\item Subdiffusive continuous time random walk (CTRW)~\cite{Scher1975}.
In the CTRW the random walker's motion is generated by a renewal process consisting of discrete jumps with a given waiting time distribution between jumps, $\psi\left(\tau\right)$, and a distance distribution of jump lengths, $f\pare{\bf{\Delta}}$. 
Within the context of this study $f$ will be a centered Gaussian distribution and $\psi\left(\tau\right)\propto \tau^{-\alpha -1}$, giving rise to subdiffusive random walks. 
The definition of the CTRW used here corresponds to a physical model associated to an annealed environment~\cite{Bouchaud1990}. 
The subdiffusive CTRW model displays anomalous property (iii) and does not have a tractable likelihood. It shows weak ergodicity breaking~\cite{Lomholt2007}, ageing~\cite{Schulz2014,Krsemann2014}, and non thermal plateau convergence when confined~\cite{Burov2010}, and it has discontinuous paths. 

\item Superdiffusive Levy walk (LW)~\cite{Klafter1994,mandelbrot1982the,Zaburdaev2015}. 
LWs belongs to the CTRW class of models, but instead of performing discontinuous jumps their motion is continuous and  composed of a series of uncorrelated ``flights" of constant speed generated from given flight time and distance distributions,  $\psi\pare{\tau}$ and $f\pare{\bm{\Delta}}$, respectively.
Here, we consider a subset of the LW class corresponding to superdiffusive motion with a flight time distribution scaling as $\psi\pare{\tau}\sim \tau^{-1-\sigma}$, 
and with a distance distribution that is conditional on $\tau$ and given by $f\pare{\bm{\Delta}} \propto \delta(|\bm{\Delta}| - \bm{v}\tau)$, where $v$ is a constant speed parameter and $\delta$ is the Dirac delta function. 
The superdiffusive LW displays anomalous property (iii), 
it exhibits weak ergodicity breaking and its likelihood is intractable.

\item Annealed transit time motion (ATTM)~\cite{Massignan2014,Akimoto2016}.
The ATTM model considers a random walker in an annealed heterogeneous environment where the diffusivity varies over space and time. In the ATTM, the random walker has a diffusivity that is piecewise constant over time with values drawn from $\pare{D}\sim D^{\sigma - 1 }$ and with resting time at each diffusivity level drawn from a conditional distribution $p(\tau|D)$.
We consider here $p(\tau|D) = \delta(\tau - D^{-\gamma})$ and the parameter range $\sigma < \gamma < \sigma + 1$ (defined as Regime I in ref~\cite{Massignan2014}), which leads to subdiffusive motion with an anomalous exponent of $\alpha = \sigma/\gamma$. 
Subdiffusive ATTM displays both anomalous properties (i) and (iii), and it exhibits weak ergodicity breaking and ageing. 
\end{enumerate}

We refer to realisations of random walks as trajectories. 
A trajectory, $\Rt = (\r_{1}, \r_2, \ldots, \r_{N})$, is a $d$-dimensional time-series of positions $\rn$ recorded at equidistant points in time $t_{n} \in \lbrace \Delta t, 2\Delta t, \ldots, N\Delta t \rbrace$. 

Experimentally, the positions of the trajectory are corrupted by various types of experimental noise. 
Here we assume an independent additive Gaussian white noise,  i.e.\ $\rnoise=\rn + \pmb{\xi}_n$, where $\pmb{\xi}_n \sim \mathcal{N}\pare{\bf{0},\sigma\mathbb{I}}$ are $d$-dimensional Gaussian variables that are independent of both the past and of the random walker's motion. 
We refer to~\cite{Vestergaard2014, Calderon2016} for how  correlated noise induced by experimental settings can be taken into account. From a statistical analysis perspective, we consider individual random walk trajectories as noisy time series with two latent variables: a continuous anomalous exponent $\alpha$ and a discrete random walk model class. 

\section{From random walks to graph learning}
\label{graphs}

We propose to leverage the flexibility and the capacity for representation learning of graph neural networks (GNNs)~\cite{1903.02428} to infer properties of random walks, in particular their anomalous exponent and model class, following a methodology which we detail below.

We start by associating a graph with each trajectory (see Figure \ref{fig:connection}). 
Inference is then performed by a GNN, capable of processing graphs of variable size, and which outputs an estimate of the anomalous exponent and a probability of belonging to each random walk model class. 
The weights of this neural network are set during training (on numerically generated trajectories) similarly to conventional supervised learning schemes, and the network can later be used for inference on other trajectories, not seen during training.

\subsection{Rationale for learning random walks with graph neural networks}
\label{w-graph}

The use of a graphical representation has long been a method of choice to model and perform inferences of complex systems~\cite{daphnekoller2009}. 
For example, factor graphs associated to mean field, belief propagation~\cite{Yedidia2011} and cavity methods~\cite{mezard2009information} have been used to model spin glass dynamics and perform complex optimisation problems. 
Hidden Markov models~\cite{bishop2006pattern} have been developed to model changes in random walker dynamics, 
and mixture models~\cite{Wainwright2007,Samuylov2019} have been applied to approximate the point spread function in fluorescence microscopy and to model and analyse complex networks~\cite{Peixoto2017}. 

Over the past 4 years, extensions of deep learning approaches to graph data in the form of graph neural networks have attracted significant attention~\cite{1903.02428} and have demonstrated great efficiency for representation learning on point clouds, graphs and manifolds~\cite{1609.02907,Qi2017PointNetDH}. 
GNNs meet several criteria that make them well suited for analyzing trajectories, and which motivated the design of our learning procedure.
(i) They can be applied directly, using a shared architecture, to trajectories of different lengths~\cite{Charles2017,Qi2017PointNetDH}.
(ii) The choice of graph structure allows taking into account different time scales while retaining a sparse architecture. 
(iii) Numerous known features associated to random walks, such as the convex hull~\cite{Lukovi2013,Lanoisele2017,Grebenkov2017}, first passage times~\cite{Redner2001,Gurin2016,Chupeau2015}, or the distributions of different features' extreme values~\cite{Mori2020,Godrche2015,Wang2020}, 
are linked to geometric properties which can be learned efficiently using GNNs.
(iv) Finally, while advances in machine learning are associated to impressive achievements~\cite{LeCun2015,1810.04805} they are often obtained with large scale models (several millions to billions of parameters) that are prone to over-fitting and are challenging to interpret~\cite{goodfellow2016deep}. 
Hence, a model with a limited number of parameters and a means to quantify the acquired information during training  would be beneficial to understand the requirements for random walk inference using machine learning. 
All these criteria point towards using graph neural networks for individual random walks analysis.

\begin{figure}[!h]
 \centering
  \includegraphics[width=.8\linewidth]{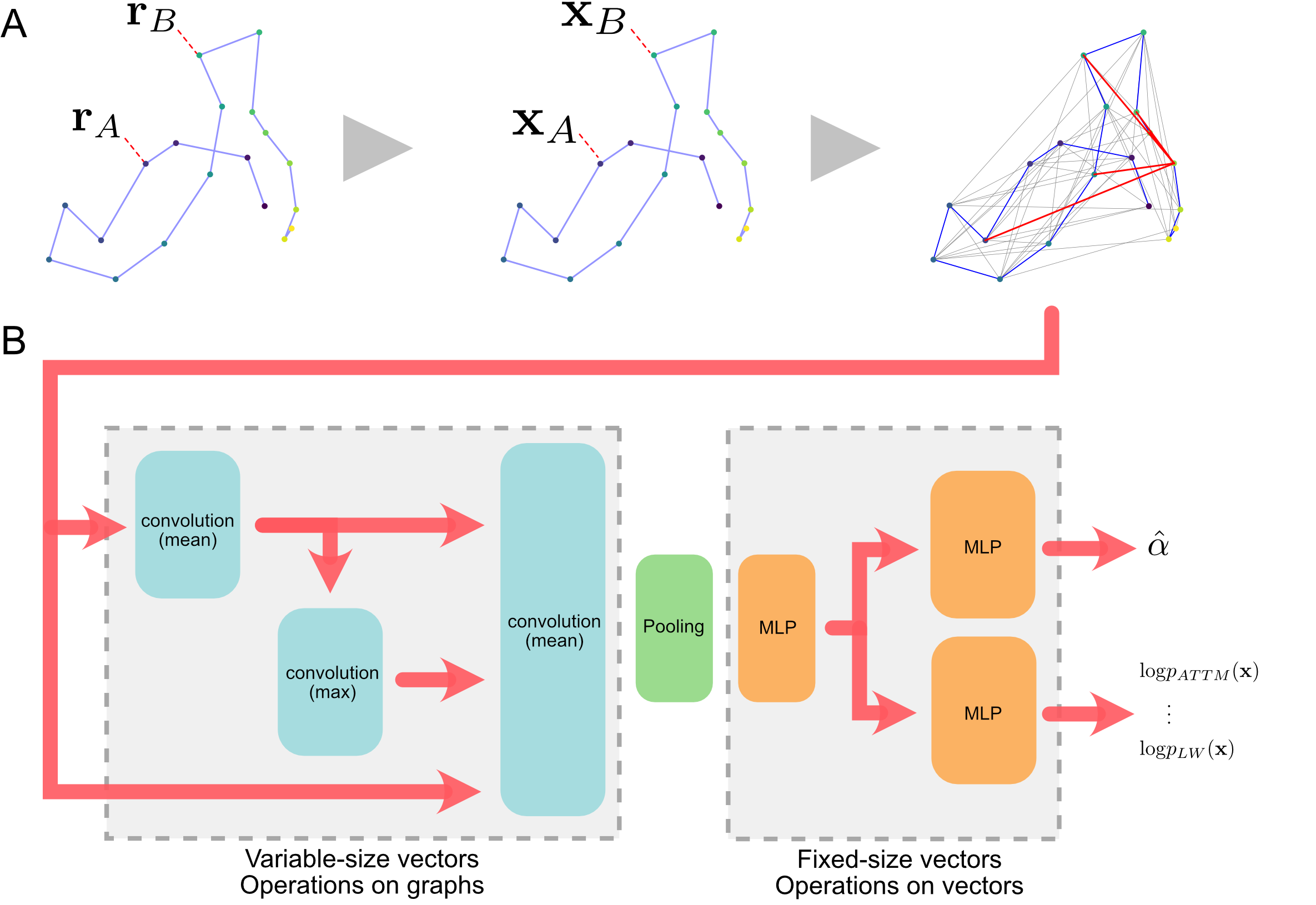}
  \caption{
  Graphical representation of the GNN model. 
  A) Construction of a graph from a raw trajectory:
  (left) raw trajectory, (middle) positions in the trajectory are represented by nodes and a vector of local features is associated to each node; (right) nodes are connected to each other using a wiring scheme as described in section~\ref{graph-traj} to form a graph on which the GNN is applied. 
  B) Overview of the GNN's architecture. 
  Red arrows symbolise outputs of layers passed as inputs to other layers.
  When several arrows point to a same layer, their features vectors are concatenated before being passed to the layer. 
  Graph convolution layers are shown in blue, they can process inputs of variable length (for graphs of variable size representing trajectories of variable length) using neural message passing (see Section~\ref{sec:conv_layers}). They use two different aggregation methods for messages: mean (averages over all the messages) and maximum (takes the maximal value of messages, for each feature). In green, the pooling operation embeds the variable-sized vector output from the graph convolution layers into a fixed-size vector representing the trajectory.
  Orange boxes downstream are multi-layer perceptrons, acting on vectors of fixed dimension as in conventional neural networks architectures.}
  \label{fig:flow}
\end{figure}

\subsection{Graph representation of trajectories}
\label{graph-traj}

We associate to each trajectory $\Rt = (\r_{1}, \r_2, \ldots, \r_{N})$ a directed graph $G = (V,E,\Xt)$, with $V = \{1, 2, \ldots, N\}$ the set of nodes corresponding to the positions in the trajectory, $E \subseteq \{(i,j) | (i,j) \in V^2\}$ the set of edges connecting pairs of nodes, and $\Xt = (\xVec_1^{(0)}, \xVec_2^{(0)}, \ldots, \xVec_N^{(0)})$ a sequence of local feature vectors $\xVec_i^{(0)}$ associated to each node $i$ in $V$ (fig.~\ref{fig:flow}A). 
Each node features vector $\xVec_i^{(0)}$, of dimension $n_x$, may contain any feature of the trajectory or of the graph and may depend only on $i$ or on arbitrary neighborhoods of $i$. 
This graph-based encoding of trajectories was inspired by applications of GNNs to point cloud data~\cite{Charles2017,Qi2017PointNetDH}, but where time is an additional feature associated to each point here.

We attach to each node $i$ the time $t_i$ and three differently normalised versions of the $i$-th position $\mathbf{r}_i$: 
(1) normalised using the standard deviation of step sizes, 
(2) normalised using the standard deviation of positions, 
and (3) normalised using the mean step size. 
We also include in each node features the values of the cumulative sums of step sizes and of squared step sizes up to their time-point, computed using each of the three normalised positions.
In order to prevent Levy walks from having a disproportionate influence in the learning process, due to the extreme distance values that the walk can induce, we clipped 
extreme jump lengths before normalisation. 
We noted during initial training that these rare events induced significant bias in the batch normalisation layers~\cite{1502.03167}.

Thus, two matrices initially represent the graph associated to the random walk: the (sparse) adjacency matrix $A$, of size $(N,N)$, and the node feature matrix $X$, of size $(N,n_x)$ where $n_x$ is the number of features initially attached t. 
Note that we may also add features to edges in the graph~\cite{1810.00826}, represented by an edge feature matrix $U$, of size $\pare{|E|,n_e}$.
Training is much faster when omitting edge features, due to the efficient implementation of sparse matrix operations. As a consequence, we do not consider GNNs leveraging edge features  in the main text, but we investigate their performance for specialised training involving limited information (see Section~\ref{sec:edge_features} in Supplementary Information) 

A known limitation of message passing GNNs has motivated us to choose particular wiring schemes for the graph. 
The mechanism of information propagation in a GNN involves iteratively passing messages between neighboring nodes, aggregating them in each step. 
The latter creates an information bottleneck~\cite{2006.05205}, leading to a limitation of information encoding in finite sized vectors. 
A GNN may fail to faithfully propagate local information stemming from nodes separated by long paths in the graph.
It can hence perform poorly if the properties to be predicted depend on long-range information,  which is generally the case for the task of classifying various random walks and inferring their anomalous exponents. 
Our approach overcomes these limitations by using structured wiring schemes to ensure more direct message passing from distant nodes.
We discuss here two different wiring schemes, (i) hierarchical causal and (ii) regular random, but many options are possible. 
In the hierarchical causal scheme (i), the incoming edges of each node connect only to nodes in the past (respecting causality): node $i$ is connected to nodes $i - \Delta_1,  \ldots, i - \Delta_{max}$, where $(\Delta_i)_{i \geq 1}$ is a geometric series (see details in \ref{supp-hyper}).
In the regular random scheme (ii), edges are drawn at random, with the only constraint that all nodes have the same \emph{in-}degree, generating a type of random regular graph. 
Example graphs can be seen in Figure~\ref{fig:connection}. 
In both schemes, the graph structure ensures that distant time points of the random walk are connected by short paths. 

\subsection{Neural network architecture}

We used a two-part architecture for the graph neural network, starting with an encoder followed by task-specific multi-layer perceptrons~\cite{goodfellow2016deep}, each estimating a property of interest from the latent representation built by the encoder -- here the anomalous exponent and the random walk class. 
This architecture, shown in figure \ref{fig:flow}, enables multi-task training (i.e.\ simultaneous inference of a random walk's class and anomalous exponent). 

The encoder is the entry-point to the model. It embeds the graph representation of the trajectory into a latent space whose dimension is independent of the trajectory length. 
To do so, it performs several graph convolution operations~\cite{1606.09375,1611.07308,1806.08804,1801.07829} (described in Section \ref{sec:conv_layers} below) which propagate learnt features through the graph. It is terminated by a pooling layer, {i.e.} an operator that combines an aggregation of features across nodes with a multi-layer perceptron that outputs a fixed-size vector. 
We used convolution layers using both ``mean" and ``max" operations to aggregate messages they receive from their neighbors, thus enabling the network to compute a broader variety of features.
We also chose to wire convolution layers so that the last one receives both the output of its predecessor and the initial features to prevent the information from vanishing through bottlenecks created by successive graph convolutions.

\subsection{Graph convolution layers and neural message passing} 
\label{sec:conv_layers}

\begin{figure}
 \centering
  \includegraphics[width=.5\linewidth]{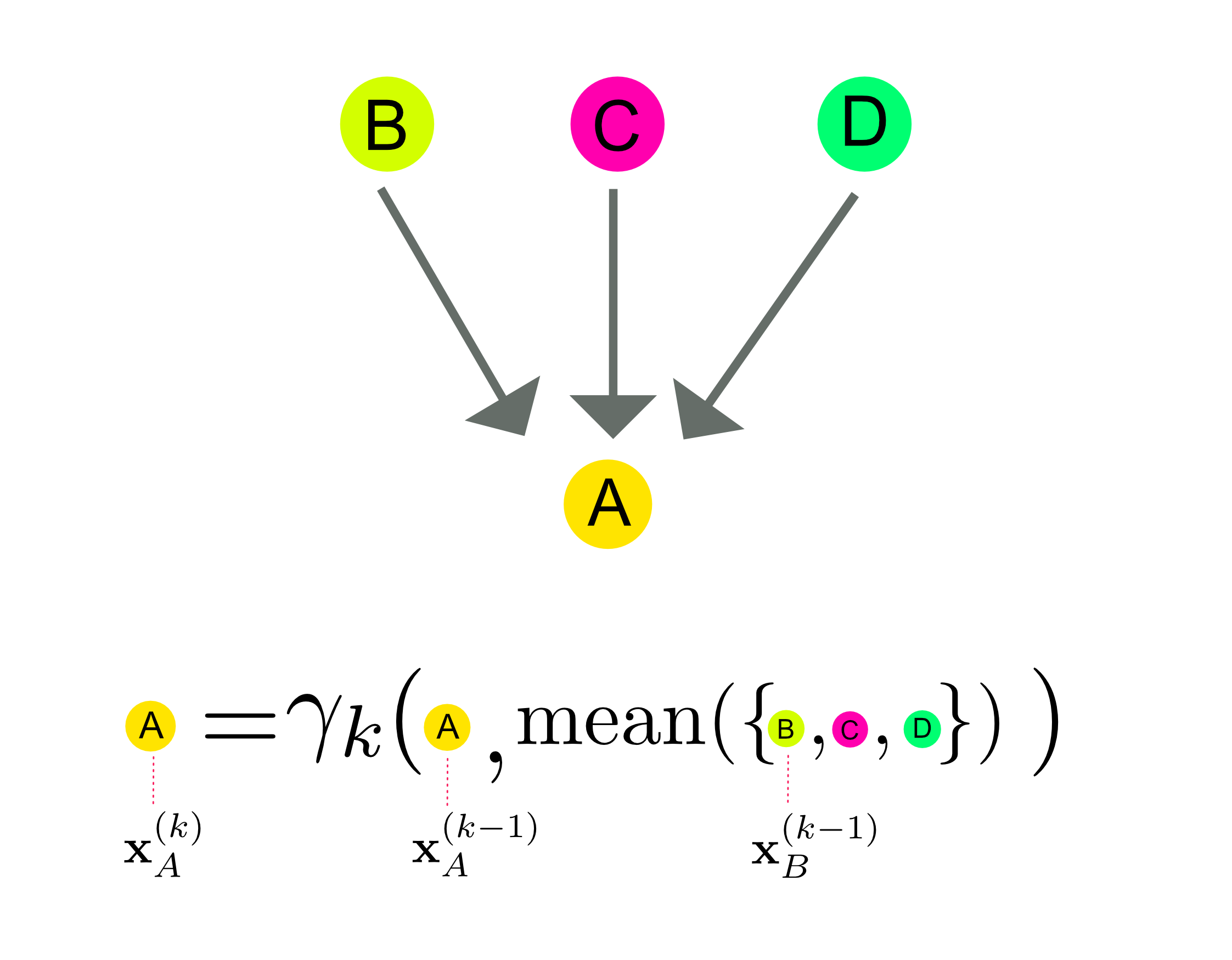}
  \caption{Illustration of a graph convolution. 
  At each iteration $k$, each node ($A$) aggregates messages sent by nodes that are connected to it.
  The color code illustrates how the information is propagated between nodes. 
  The weight parameters of the multi-layer perceptron $\gamma_k$, corresponding to the $k$th layer of the GNN, are learnt during the training. Here, the aggregation scheme for feature vectors shown is a mean over neighbor nodes.}
  \label{fig:graph_conv}
  
\end{figure}

The core of GNN operations is formulated in terms of neural message passing, which gathers and transmits information from nodes to nodes through connecting edges (Fig.~\ref{fig:graph_conv}) and aggregates it using basic operations such as convolution and pooling
\cite{1606.09375,1611.07308,1806.08804,1801.07829,Hamilton2020}. 

Graph convolution layers implement operations on node feature vectors following a message passing scheme~\cite{Hamilton2020} (Fig.~\ref{fig:graph_conv}):
\begin{equation}
\label{update_formula_simple}
\vec{x}^{(k)}_{i} \leftarrow \gamma_k\pare{\vec{x}^{(k-1)}_{i}, \mathcal{X}_{j\in \mathcal{N}\pare{i}}\vec{x}^{(k-1)}_{i}} \enspace.
\end{equation}
Here, the exponent $(k-1)$ denotes a vector's value {before} the $k$-th convolution, and $(k)$ its updated value after the convolution. 
The functions $\gamma_k$ are neural networks (in our case, multi-layer perceptrons) whose weights are learned during training. 
The dimension of $\vec{x}^{(k)}_{i}$ is not constrained to be the same as that of $\vec{x}^{(k-1)}_{j}$: drawing analogy with convolution layers in classic architectures used for image processing, each output dimension of $\gamma$ corresponds to a convolution kernel. 
$\chi$ is a permutation-invariant aggregation operator which reduces the set of input vectors to a single vector of fixed dimension. 
In our case, it is either a feature-wise mean or maximum across nodes. We illustrate both the graph building process and the learning in Figure~\ref{fig:flow}. 

We point out here that, while initial motivations for using GNNs stemmed from the known efficiency (see Section~\ref{w-graph}) of graph models in physics, GNNs differ strongly from physically motivated graph models and message passing techniques. 
The neural messages do not represent beliefs about features or variables of interest, they are not normalised by conservation of probability and optimisation is not performed by sampling. The graph serves as a means to link features to sets of neighbor features and allow "classical" learning to be performed by optimisation.

\section{Results}

We test our model's performance on classification (model selection) and regression (parameter estimation) tasks for simulated trajectories of varying lengths and with a range of localisation noise amplitudes and characteristic scales of motion. 
We use the same convention as in the AnDi Challenge~\cite{andi2020} regarding the localisation noise, i.e.\ we apply an independent Gaussian noise to all positions in a trajectory with a standard deviation equal to a constant factor of the expected standard deviation of the jump sizes of the random walk. 
We refer to this proportionality factor as the ``noise amplitude", and we consider noise amplitudes in the range $[0,1]$.

The lengths and anomalous exponents of trajectories used for both training and evaluation were sampled uniformly between their respective extreme values (unless otherwise specified, $N=10$--$1000$ and $\alpha=0.05$--$1.95$). 
For each evaluation, performance metrics were computed using one million trajectories (200\,000 from each model), generated using the AnDi package~\cite{andi2020}. 
We added localisation errors with the same amplitude to all dimensions of a trajectory. 
Here, we will show only results for 3D trajectories, but the dimension of the trajectories does not qualitatively change results. 

Unless otherwise specified, we trained the GNN to perform both classification and regression simultaneously by using a training objective given by a simple sum of the mean squared error (MSE) of the estimated anomalous exponent and the cross-entropy between the true and predicted class labels. 
To quantify regression accuracy, we follow the AnDi settings~\cite{andi2020} and use the mean absolute error (MAE) between the estimated and true values of the anomalous exponent $\alpha$, while we use the $F_1$ score to quantify overall classification accuracy and confusion matrices for class-by-class evaluation.
Section~\ref{supp-hyper} gives detailed definitions of each of the measures.

\label{results}
\subsection{Performance in absence of localisation noise}

\begin{figure}
 \centering
  \includegraphics[width=.95\linewidth]{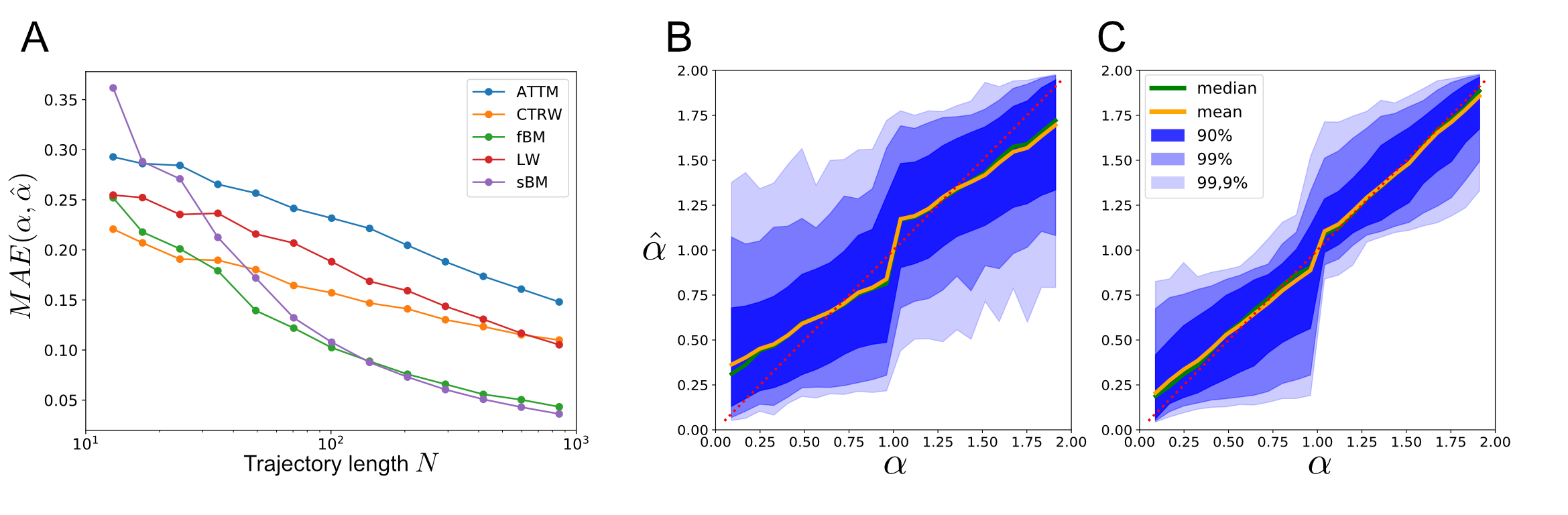}
  \caption{ Performance of the inference of the anomalous exponent $\alpha$ on 3D trajectories. 
  A) Mean absolute error (MAE) of estimate of the anomalous exponent, $\hat\alpha$, as a function of trajectory length $N$.
  B) \& C) 
  Distribution of $\hat\alpha$ values as a function of true value of $\alpha$ for B) trajectories of 10 to 100 points and C) 100 to 1000 points. $x$-axis : true exponent $\alpha$, $y$-axis : inferred exponent $\hat{\alpha}$.}
  \label{fig:alpha_no_noise}
  
\end{figure}

\begin{figure}[!h]
 \centering
  \includegraphics[width=.7\linewidth]{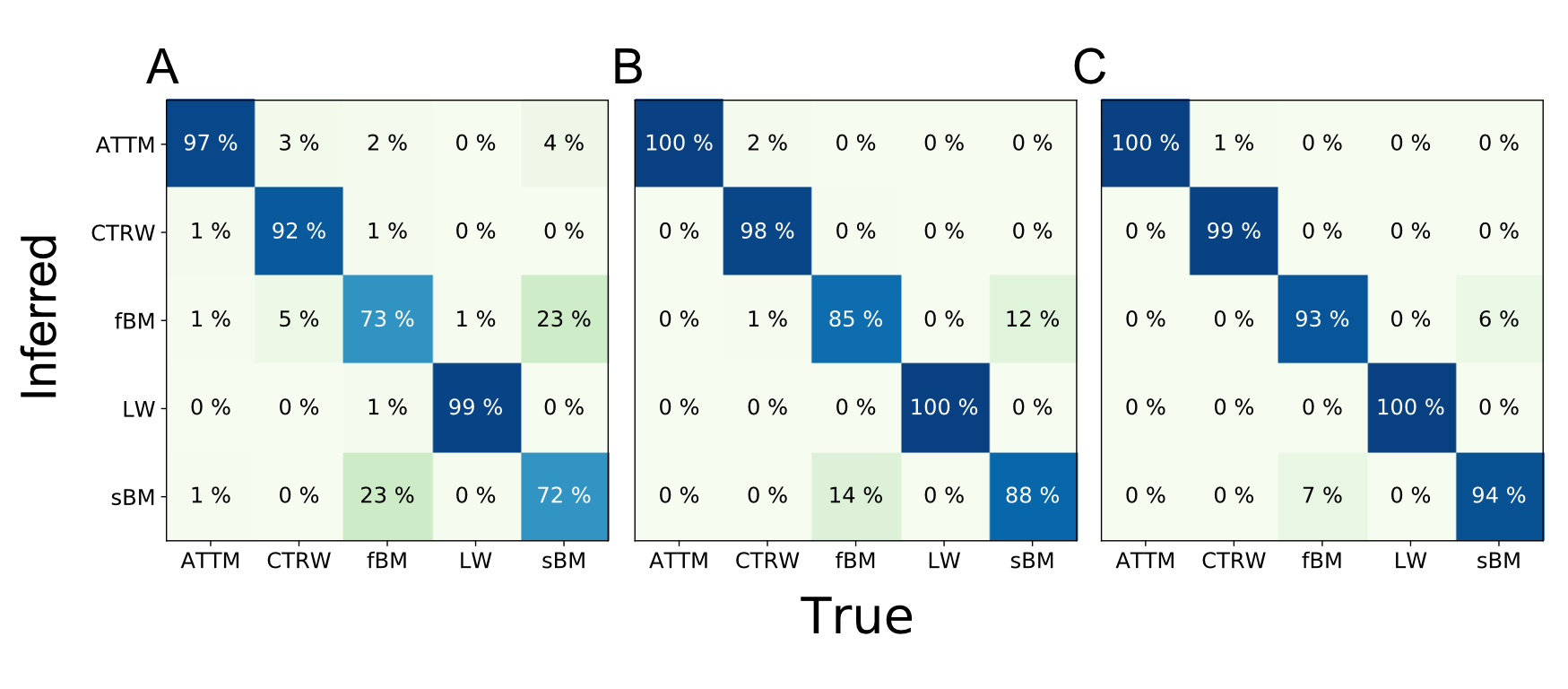}
  
  \caption{Confusion matrix for model classification, 
  {i.e.} probability to identify a trajectory as having been generated by each random walk model (inferred class) given its true generating model (true class). 
  For trajectory lengths in the range: A) 10-50, B) 50-200, and C) 200-1000.}
  \label{fig:model_no_noise}
  
\end{figure}

We show in Figures \ref{fig:alpha_no_noise} and \ref{fig:model_no_noise} the performance of a GNN trained on trajectories of lengths  between 10 and 1000 and in the absence of localisation noise. 
Figure~\ref{fig:alpha_no_noise}A shows that the accuracy of the inference of the anomalous exponent $\alpha$ from a single trajectory depends on trajectory length and the class of random walk considered: the anomalous exponent of ATTM is harder to infer than that of CTRWs or fBMs. Looking at how the estimation error on $\alpha$ depends on its true value, we see as can be expected a conservative bias shifting estimates away from extreme $\alpha$ values (i.e.\ 0, 1 and 2).
The bias is pronounced for short trajectories and decreases with trajectory length (Fig.~\ref{fig:alpha_no_noise}B,C). 
The bias stems in majority from the poor performance on ATTM trajectories.

Looking at the confusion matrices to assess classification accuracy (Fig. \ref{fig:model_no_noise}), we see that even for short trajectories, most walks are accurately classified, and misclassifications mainly confuse sBM and fBM. 
For trajectories longer than 200 points, the classification exhibits high performance. 
The GNNs classification performance for short trajectories illustrates that it relies not only on the asymptotic properties  but also on finite-scale features of the random walks. 
Hence, even for short trajectories identification is possible. 
Furthermore, the confusion between sBM and fBM can be explained by the similarity of the processes for short trajectories, with a more pronounced effect for $\alpha$  close to one (in the range of approximately 0.7 to 1.4). 
sBM and fBM indeed share the same marginal probability density for the time-evolution of the walker's position~\cite{Lim2002} and they both approach Brownian motion as $\alpha$ approaches one.

\subsection{Robustness to noise}
\label{robustness}

Experimentally recorded trajectories are subject to various sources of noise. 
We here focus only on localisation noise, modelled as an uncorrelated Gaussian noise, but correlated noise sources may also be present~\cite{Berglund2010,Vestergaard2018}. 
We investigated the performance of the inference when both training and inferring on trajectories observed with a broad range of noise amplitudes. 

\begin{figure}
 \centering
  \includegraphics[width=.70\linewidth]{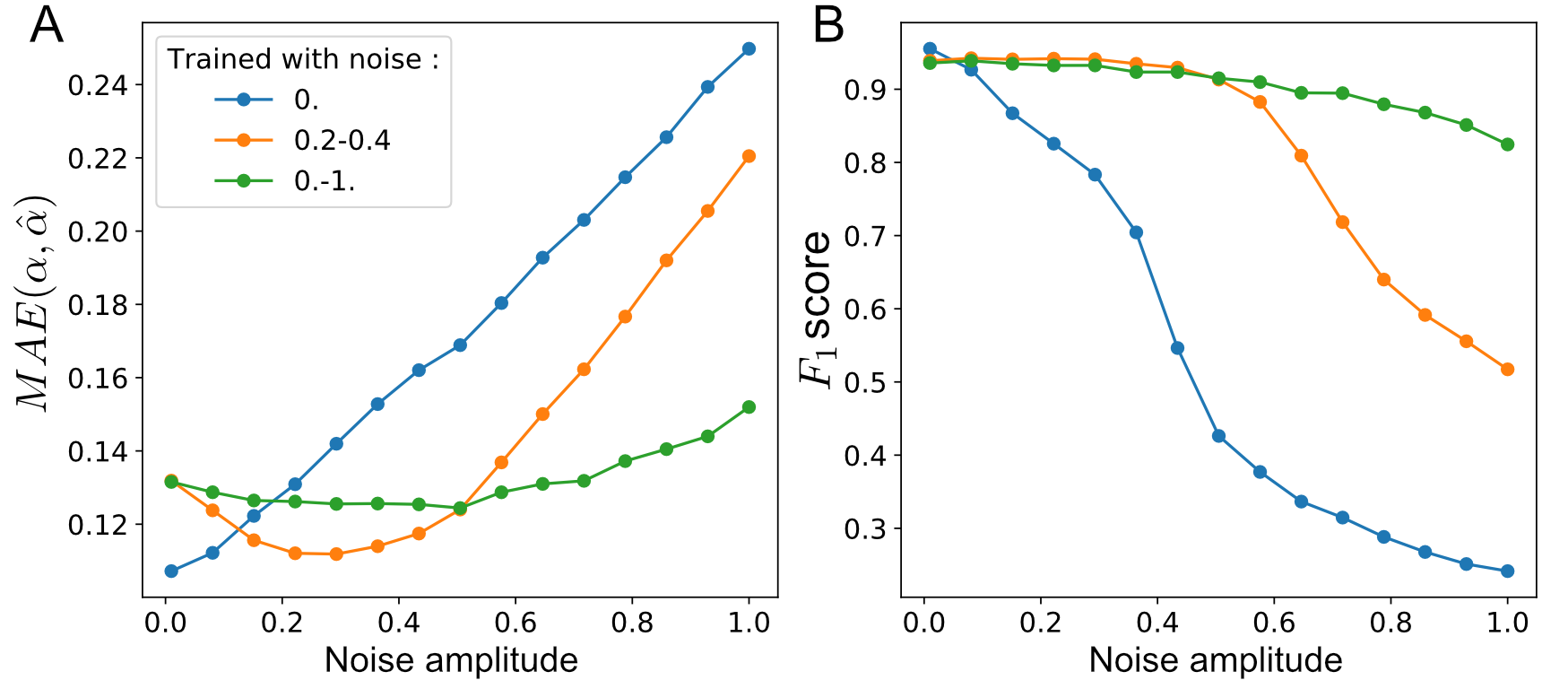}
  \caption{Inference performance as a function of noise amplitude and for different ranges of noise amplitudes included in the training data. 
  A) Mean absolute error of the anomalous exponent inference, $\mathrm{MAE}\pare{\hat\alpha}$.
B) $F_1$ score for model identification. The $F_1$ score is the harmonic mean of precision and recall (see details in Supplementary Section~\ref{supp-hyper}).}
  \label{fig:noise}
\end{figure}

As shown in Figure~\ref{fig:alpha_no_noise}, high localisation noise may significantly impair the accuracy of inferences when it has not been taken into account in the training data. 
We tested a global approach to induce noise robustness by assuming that no information on noise was accessible safe for a range of possible amplitudes. Hence, we trained on trajectories with added localisation error whose noise amplitude was randomly drawn from $[0,1]$. 
Remarkably, training on such a wide distribution of noise  leads to a nearly flat performance in of the anomalous exponent inference over the full range of noise amplitudes (green curve in Fig.~\ref{fig:noise}). 
Similarly, the performance in classifying the trajectories, i.e.\ identifying their generative model, exhibited only a limited decrease with increasing noise amplitude. 
Conversely, when inference is performed on trajectories with localisation noise outside the range that the GNN was trained on, the performance of both regression and classification may degrade significantly.

In most experimental settings, there are means by which a range of possible values for positioning noise can be either deduced \textit{a priori} or directly measured. 
Taking this in account by simulating with the same noise range to generate the training set increases the accuracy (curve corresponding to 0.2-0.4 in Fig.~\ref{fig:noise}A), even if the GNN trained on the whole range of noise amplitudes already performs well. 
We refer the reader to~\cite{Lindn2017} for an example of a procedure to estimate positioning noise within the context of single molecule experiments.

\subsection{Improving performance on specific cases}

\begin{figure}
 \centering
 \includegraphics[width=.8\linewidth]{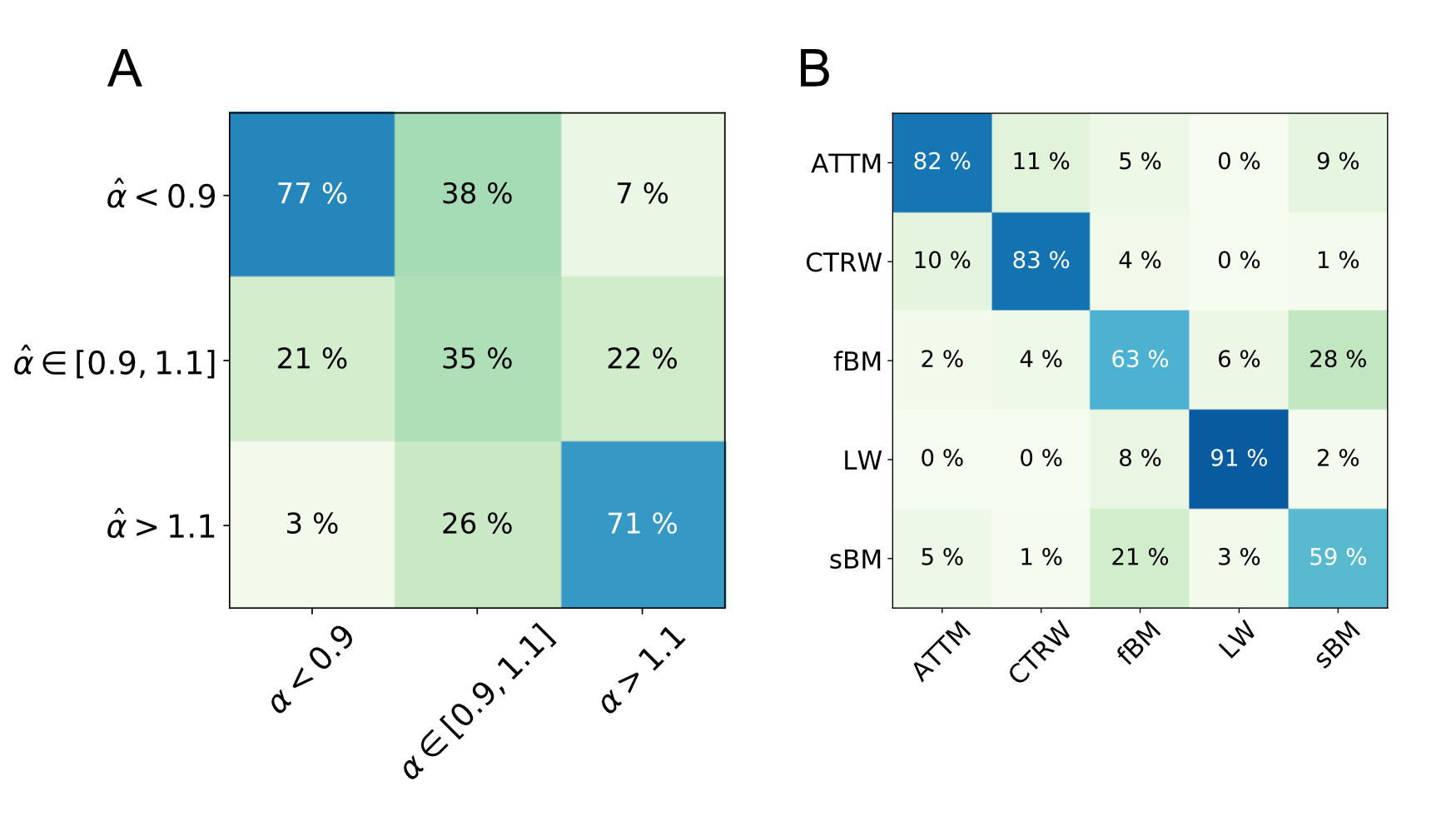}
 \caption{Performance on short trajectories ($10 \geq N \geq 50$) for a GNN trained with a known range of localisation noise amplitudes. 
 A) Confusion matrix for identification of subdiffusive, normally diffusive and superdiffusive anomalous exponent values. 
 B) Confusion matrix for model classification. 
 The color code is identical to the one in figure~\ref{fig:model_no_noise}.}
 \label{fig:SPT}
\end{figure}

We tested the performance of the architecture on data with parameters corresponding to typical experimental conditions, {i.e.} short trajectories (between 10 and 50 points) corrupted by a significant but bounded amount of noise (here, noise amplitudes between 0.2 and 0.4). The model 
was trained on this same range of trajectory lengths and noise amplitudes. 
We illustrate the model's performance in Figure~\ref{fig:SPT}. We show that despite the inherent difficulty of inferring properties from such short observations, the model is precise enough to extract relevant properties from these trajectories, {e.g.}, it can reliably separate subdiffusive and superdiffusive trajectories and distinguish CTRWs from fBM. 
These results suggest that if we consider the random walk models to provide good approximations of the dynamics we may encounter in an experimental system, the approach may be efficiently leveraged to analyse single molecule experiments.

\begin{figure}
 \centering
 \includegraphics[width=.85\linewidth]{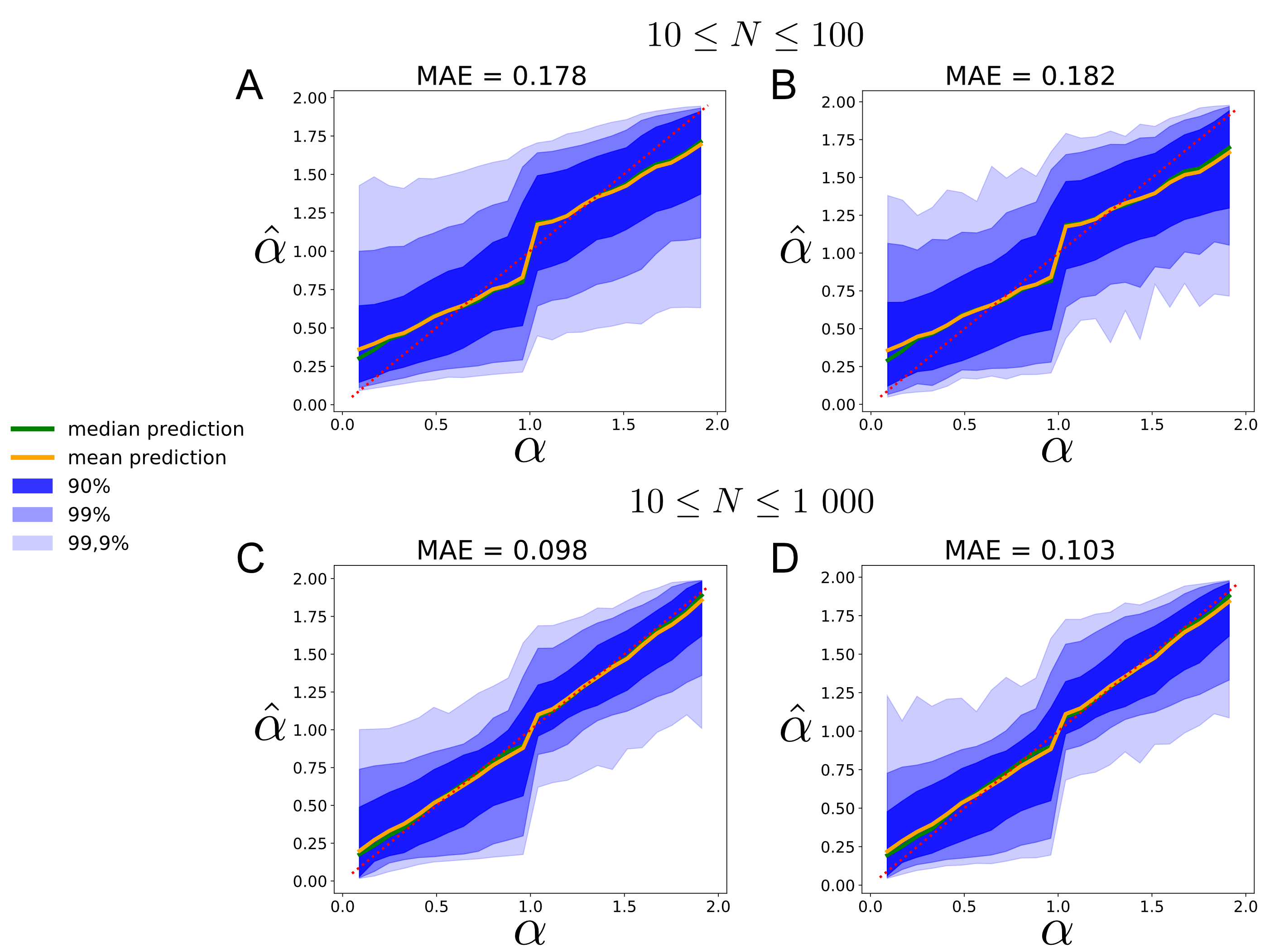}
 \caption{ Improvement in model performance with specialized training: 
 Distribution of estimated anomalous exponent values, $\hat\alpha$ as a function of the true value $\alpha$ for GNNs trained on only short or both short and long trajectories when applied to analyse short trajectories (A,B) and for GNNs trained only for regression or for both regression and classification (C,D).
 A) GNN trained only on short trajectories, $10 \leq N \leq 100$, and B) GNN trained on both short and long trajectories, $10 \leq N \leq 1000$, both applied to short trajectories of lengths $10 \leq N \leq 100$. 
 C) GNN trained only on the anomalous exponent estimation task and D) GNN trained on both tasks.}
 \label{fig:specific}
\end{figure}

We next assessed more generally the effect of specialisation of the inference task on performance. 
First, we compared the performance of a model trained specifically on short trajectories ($10 \leq N \leq 100$) to that of a model trained on short and long trajectories ($10 \leq N \leq 1,000$)  (Fig.~\ref{fig:specific}A,B).
We only considered model performance on the range of lengths that was common to both training sets. 
Then, we compared the performance of a model trained solely to infer the anomalous exponent with one trained both for regression and classification (Fig.~\ref{fig:specific}C,D).
We see that the performance increases with specialized training, although here only to a limited  extent. 
This capacity of GNN models trained on numerical simulation to provide generalisable inference will be instrumental for their use in experimental data analysis.

\subsection{Performance as a function of the number of neural network parameters}

\begin{figure}
 \centering
 \includegraphics[width=.5\linewidth]{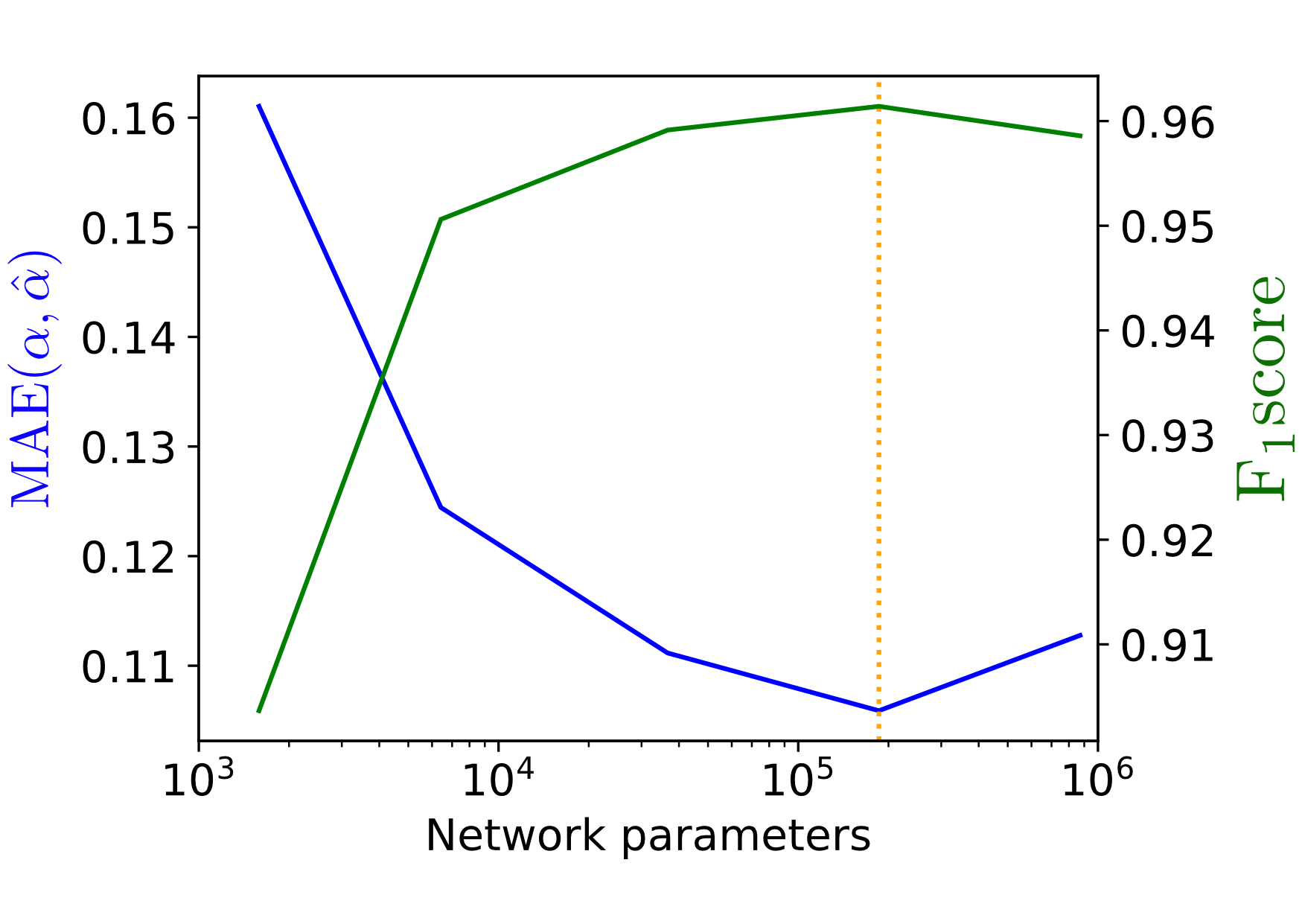}
 \caption{ Inference performance (MAE and $F_1$ score) as a function of the number of neural network parameters. The vertical yellow line indicates the number of parameters used in the rest of the paper.}
 \label{fig:ParamsPerf}
\end{figure}

GNNs generally do not require as many parameters as other recent deep neural architectures~\cite{1609.02907}. 
Here, we investigated the effect of the GNN's size (as measured in total number of parameters) on its performance. 
To do so, we modified the base architecture, first by making layers thinner and then by removing some layers (reducing by up to two orders of magnitude the number of parameters). Details are available in Table~\ref{tab:params}.

Results are shown in Figure~\ref{fig:ParamsPerf}. 
It is noteworthy that even with only about 1,600 parameters, the GNN maintains a good performance. This suggests that increased model tractability may be possible for GNNs.

\section{Discussion}
\label{discussion}

\subsection{Latent space encoding of physical properties and generalisation}

Our work lies in the framework of simulation-based inference~\cite{Cranmer2020}. 
This allowed large scale data generation at low computational cost and ensured that features of one dataset could not impair the learning. This contributed to reducing possible overfitting bias~\cite{mostafa}. However, since some processes are non-ergodic, it is still a concern that the learning procedure might have failed to learn relevant properties on a finite dataset. Moreover, the statistics of experimentally recorded trajectories are unlikely to exactly match those of any of the models that we trained our machine learning model on. 
Here, we show that the learned latent space of the GNN encodes physically relevant properties of random walks, and we explore how it can be used to evaluate the robustness and generalisation performance of the approach.

\begin{figure}[!h]
 \centering
 \includegraphics[width=.95\linewidth]{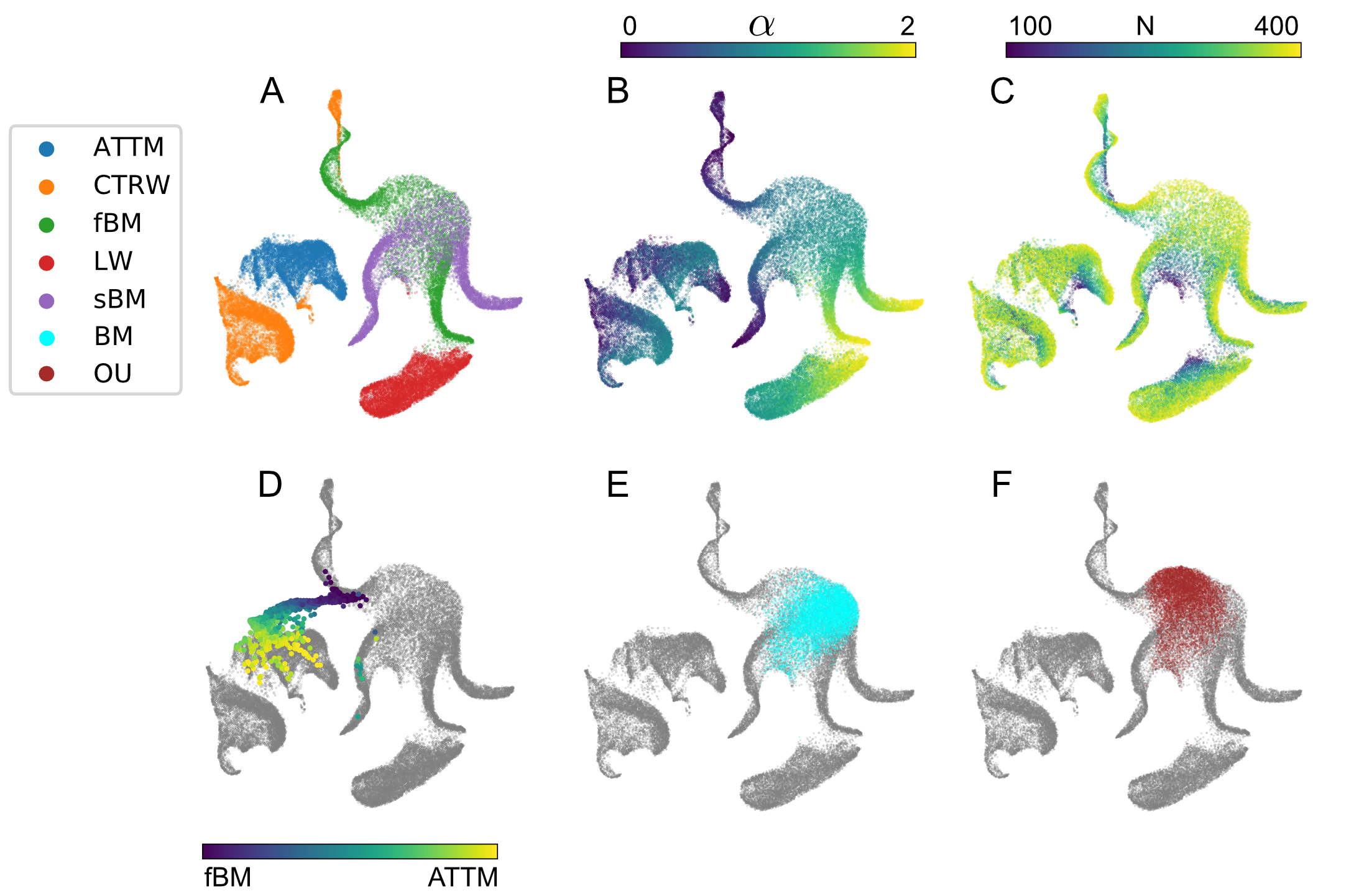}
 \caption{UMAP~\cite{1802.03426} representation of the latent space of the GNN. UMAP is a generalist  manifold learning and dimension reduction algorithm relying on  Riemannian geometry and algebraic topology to perform the embedding. Each point represents the encoded latent position of a single trajectory.
 A) Positions of each trajectory colored by the random walk's true model class.
 B) Mapping of the anomalous exponent $\alpha$ in the latent space. The color is associated with the value of $\alpha$. 
 C) Mapping of the trajectory length in the latent space. The color is associated with trajectory length.
 D) Positions of trajectories mixing fBM and ATTM in different proportions. Colors indicate the percentage of the trajectories generated by each of the two random walks. Note that trajectories continuously occupy the empty space between the ATTM cluster and the fBM domain. 
 E) Projection of Brownian trajectories in the latent space. 
 F) Projection of Ornstein-Uhlenbeck trajectories in the latent space.  
 }
 \label{fig:latent}
\end{figure}

In Figure~\ref{fig:latent} we provide a representation of the latent space (output of the penultimate MLP module). 
Each point corresponds to the latent representation of a single trajectory.
We relied on UMAP~\cite{1802.03426}, a dimensionality reduction algorithm in the family of  manifold learning techniques, for 2D visualisation of the high dimensional latent space while preserving its local topology.

First, the structure reveals how well the model is able to distinguish trajectories of different types and anomalous exponents. 
Levy Walks, CTRWs and ATTM form three well separated clusters while fBMs and sBMs form a more complex shape with an extended overlapping region corresponding to $\alpha \approx 1$, i.e.\ the regime of normal (Brownian) diffusion (Fig.~\ref{fig:latent}A).
Within each cluster and continuous region, the anomalous exponent is mapped in a stable fashion (Fig.~\ref{fig:latent}B).
In concordance with the lower performance on regression for the ATTM model, we see that the value of $\alpha$ is less clearly mapped than for the other random walk models. 
The length of the trajectories, which directly relates to the available quantity of information, is also encoded in an ordered manner within the latent space, along a direction that is roughly orthogonal to the direction encoding the exponent.

In order to investigate the capacity of the GNN to encode physical properties of the random walks and its ability to generalize to trajectories with unseen properties, we applied the GNN to trajectories of walkers starting with a given type of motion and ending with another, and to random walks generated by unseen models on which the GNN was not trained. 

To evaluate how the GNN encodes random walks that change of motion class over time,  we generated trajectories of fixed length ($L = 200$) where the first part was generated by a given model of subdiffusive walks and the second by another, and varied the relative importance of the two parts. Both segments have $\alpha = 0.5$.
In Figure \ref{fig:latent}D we 
can see the trajectories' encoded positions in the latent space draw a transition from the fBM domain to the ATTM cluster as the percentage of the ATTM part in the trajectory increases. Some trajectories fall within a region of the latent space that was originally not occupied. The model is thus able to continuously encode random walk properties and interpolate between the properties of the two models based on previously unseen behaviour. It is an indication of its ability to generalize.

To investigate how the GNN behaves when used on trajectories generated by models not included in the training phase, we use the GNN to encode the properties of trajectories generated by pure Brownian motion (BM, with $\alpha=1$), and by the Ornstein-Uhlenbeck (OU) process which models Brownian motion confined in a harmonic potential~\cite{Gardiner}. 
In Figure~\ref{fig:latent}E we show that BM trajectories all fall within the portion of the latent space where fBM and sBM overlap, corresponding to the region where $\alpha = 1$ and their dynamics approach Brownian motion. 
The OU trajectories cover a region where  subdiffusive fBMs with $\alpha$ values close to, or slightly below one are encoded. 
This is also a physically sensible encoding as the OU process shows anticorrelated dynamics, similar to the fBM but with a much faster, exponentially decaying kernel. 
In this respect, the latent space does encode relevant physical properties, suggesting that the GNN will generalize well to experimental data with statistical properties that may not be exactly equal to any of the random walk models it was trained on.

\subsection{Misclassified random walks}

\begin{figure}
 \centering
 \includegraphics[width=.9\linewidth]{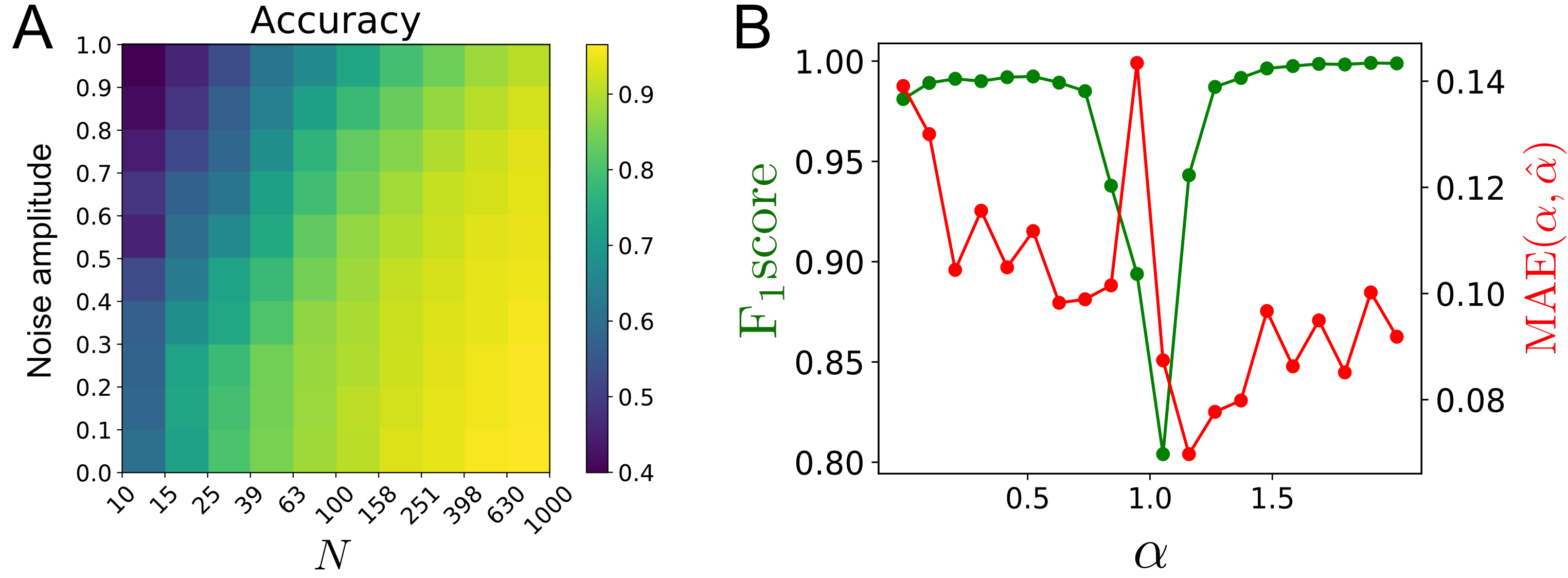}
 \caption{A) Accuracy of random walk classification as a function of trajectory length and noise amplitude. 
 B) Classification accuracy and MAE of the estimated anomalous exponent $\hat{\alpha}$ as a function of the true value of the anomalous exponent. Trajectories are of length $10 \leq N \leq 1~000$, and without positioning noise.}
 \label{fig:accuracy}
\end{figure}

We investigated the sources of misclassification by the GNN. 
In Figure~\ref{fig:accuracy}A we show the classification accuracy as a function of both trajectory length and noise factor. Error in model identification is concentrated in high noise and low length regions.
Figure~\ref{fig:accuracy} helps to pinpoint the hardest samples to characterise. Intuitively, they are short and noisy. 
Furthermore, as could be expected from the latent space's structure, properties of trajectories whose anomalous exponent is close to one tend to be harder to infer (Fig.~\ref{fig:accuracy}B).

\subsection{Influence of graph structure}
\label{graph-structure}n
iere, we investigated GNNs applied to graphs generated from recorded trajectories using a specific wiring scheme. 
As discussed in Section~\ref{graph-traj}, we designed this scheme to ensure that the graph would connect distant portions of the trajectories, for efficient information exchange, while retaining a causal and hierarchical dependency structure. To test the influence of graph structure, we compare its performance to that of a random wiring scheme in Supplementary Section~\ref{random_vs_uniform}. 

Recent work~\cite{alon2020bottleneck} has highlighted a strong link between the how a GNN's depth affects its performance and the connectivity of the graphs it is applied to. 
Following this direction, it has been proposed to learn the graph structure itself from an input point cloud in Euclidean space ~\cite{kazi2020differentiable}. 
This could very well be applied to this setting in the future.

\subsection{Computational Complexity}

The inference-time algorithmic complexity of the GNN model depends on the three main time-consuming parts when applying it to infer the properties of a random walk:
\begin{enumerate}
    \item initial features evaluation;
    \item forward pass through graph convolutions;
    \item forward pass through subsequent layers.
\end{enumerate}

1. The set of features we use to initialize each node's feature vectors can be computed in $O(N)$ time ($N$ is the number of nodes, equal to the trajectory length). 

2. From Eq.~(\ref{update_formula_simple}) it follows that the complexity of a convolution operation for all nodes of the graph is of $O(|E|)$ time complexity in the general case. In our case we restrict node's degrees to a maximum of $k$, so we have $E \leq k\, N$ and the convolution operation thus has $O(N)$ time complexity too.

3. Finally, as the dimension of the latent representation is independent of the graph's size, the forward pass through subsequent layers has $O(1)$ time complexity.

Thus, provided that the number of edges scales linearly with the number of nodes, this architecture can scale well to long trajectories, inferring their model class and anomalous exponent in $O(N)$ time.

\section{Conclusion}

In this paper we have shown that we could learn a physical representation of anomalous random walks using GNNs. We leveraged this representation to infer both the anomalous exponent and the model of a random walk from single trajectories. We relied on
simulations to train our procedure. The scheme was found to be efficient in performing regression and classification tasks as well as being robust to positional noise. We showed that the latent space learned by the GNN is linked to the physical properties of the random walks. 

While GNNs provide a general and expressive framework, they are still a new approach. Future developments are likely to improve their computational efficiency (e.g., improving their scalability to large graphs~\cite{1902.10130} and the efficiency of GPU acceleration~\cite{1902.10130}), their statistical power (e.g., by incorporating higher-order geometric features~\cite{Hamilton2020, Gainza2019} or appropriately relaxing the permutation-invariance of the aggregation operator), as well as our theoretical understanding of their capacities~\cite{1810.00826}. 

Representation learning  paves the way to new approaches to explore biomolecule random walks whose dynamics cannot be purely described by a unique canonical random walk model. 
We foresee two directions for developments: 
First, neural networks and feature learning may be used to accelerate likelihood-free inference~\cite{Cranmer2020}, allowing to fit more complex and realistic simulation-based models for experimentally recorded random walks. 
Second, the ability to learn relevant representations from random walk realisations may be exploited to develop unsupervised approaches to analyse experimentally recorded random walks.

Future work may also involve imposing constrains during learning to reinforce known symmetries in the random walk (e.g.\ directional symmetries) to increase training efficiency and to ensure that the neural network does not learn spurious features.

\vskip 0.3in
\textbf{Acknowledgments.}
We thank the organizers of the AnDi challenge, which motivated this work. We thank Thomas Blanc, Mohamed El Beheiry, Srini Turaga, Hugues Berry, Raphael Voituriez $\&$ Bassam Hajj for helpful discussions.
This study was funded by the Institut Pasteur, \emph{L'Agence Nationale de la Recherche}~(TRamWAy, ANR-17-CE23-0016), the INCEPTION project (PIA/ANR-16-CONV-0005, OG), and the \emph{``Investissements d'avenir"} programme under the management of Agence Nationale de la Recherche, reference ANR-19-P3IA-0001 (PRAIRIE 3IA Institute).  

The funding sources had no role in study design, data collection and analysis, decision to publish, or preparation of the manuscript.
\textbf{Conflicts of interest.}
The authors declare to have no financial or non-financial conflicts of interest.

\section{Supplementary Material}

\subsection{Numerical simulations, Graph neural network training and hyper parameters}
\label{supp-hyper}

\paragraph{Simulating trajectories}

Random walks were generated using the python package provided during the AnDi challenge~\cite{andi2020}. We slightly modified it to generate the same noise amplitude along all dimensions, which corresponds more closely to experimental conditions.
    
We additionally considered pure Brownian motion and the Ornstein-Uhlenbeck process: 
\begin{itemize}
  \item We simulated Brownian motion by directly sampling its displacements according to $\bold{X}_{n+1} = \bold{X}_n + \pmb{\Delta x_n}$ with $\pmb{\Delta x_n} \sim \mathcal{N}(\pmb{0},\mathbb{I})$.
  \item We simulated the Ornstein–Uhlenbeck process using the Euler method according to the following update formula: $\bold{X}_{n+1} = \bold{X}_n (1-\delta t) + \sqrt{\delta t}\pmb{\epsilon}$ with $\pmb{\epsilon} \sim \mathcal{N}(\pmb{0},0.1\mathbb{I})$ and $\delta t = 0.01$.
\end{itemize}

\paragraph{Neural network hyper-parameters}

The detailed architectures of the model's components are summarized in table \ref{tab:params}, ordered from the smallest to the largest tested network in terms of the number of parameters. Within this paper we have mostly discussed results associated to the architecture showing the best performance (see vertical line in~\ref{fig:ParamsPerf}), which correspond to the fourth row in the table. The first convolution layer receives the initial nodes features (there are 28) as well as, optionally, some of their powers. In the first two architectures presented in table \ref{tab:params}, it receives only the first power. In the third and fourth, it additionally receives the squared features. In the last architecture it also receives the cubed features (hence the initial width being a multiple of 28). This is meant to allow the network to compute moments of the features distributions. When building the graphs, we used a maximal $in-$degree value of 20.

\begin{center}
\begin{tabular}{ |>{\rowmac}c|>{\rowmac}p{.23\textwidth}|>{\rowmac}c|>{\rowmac}c|>{\rowmac}c<{\clearrow}| }

 \hline
 parameters & $\gamma$ layers & projector & $\alpha$ module & classifier\\
 \hline
 \hline
 1~588 & (56,8) \par (16,8) \par (32,8)  & (24,6) & (6,16,1) & (6,5) \\ 
  \hline
 6~420 & (56,16,16) \par (32,16,16) \par (64,16,16)  & (48,8) & (8,64,16,1) & (8,16,5) \\ 
 \hline
 36~660 & (84,32,32,32) \par (64,32,32,32) \par (128,32,32,32)  & (96,64,16) & (16,128,64,16,1) & (16,16,16,5) \\ 
 \hline
 \setrow{\bfseries}
 185~879 & 
 (84,128,64,64) 
 \par (128,128,64,64) 
 \par (256,128,64,64)  & (192,128,64,32) & (32,128,128,64,16,1) & (32,64,32,5) \\ 
 \hline
 871~596 
 & (112,256,128,128,128) \par
 (256,256,128,128,128) \par 
 (512,256,128,128,128)
 & (384,512,256,128,64) & (64,128,128,128,64,1) & (64, 128,64,32,5) \\ 
 \hline
\end{tabular}
\label{tab:params}
\end{center}

We used batches of 128 trajectories for training, with a learning rate of $10^{-3}$, exponentially decaying until it reaches $2\cdot10^{-4}$ after the network has seen $3\cdot10^6$ trajectories. Training lasts about 5 to 10 hours. We relied on the "PyTorch Geometric" package~\cite{Fey_Lenssen_2019} to implement the graph convolutions and perform the learning. 

\paragraph{Metrics}

The mean square error (MSE) is computed as follows: $\mathrm{MSE}\pare{\hat\alpha} = \langle (\hat{\alpha} - \alpha)^2\rangle$.

The mean absolute error (MAE) reads: $\mathrm{MAE}\pare{\hat\alpha} = \langle |\hat{\alpha} - \alpha|\rangle$

The cross-entropy loss, used for the classification task, reads: $\mathrm{CE} = -\langle \sum_{i =1}^{5}\delta_{m,i}\log \hat{p}_i\rangle$. Where $m \in \lbrace{1,2,3,4,5\rbrace}$ denotes the index of the true model of a trajectory.

To quantify overall classification performance, we used the $F_1$ score, which is the harmonic mean of the \emph{precision} and the \emph{recall} of the model. Using TP, FP $\&$ FN to denote the number of true positives, false positives and false negatives, respectively, we have:
\begin{itemize}
    \item Precision: $P = \frac{\mathrm{TP}}{\mathrm{TP} + \mathrm{FP}} $
    \item Recall: $R = \frac{\mathrm{TP}}{\mathrm{TP} + \mathrm{FN}}$
    \item $F_1$ score: $F_1 = 2\frac{\mathrm{PR}}{\mathrm{P}+\mathrm{R}}$
\end{itemize}

The confusion matrices, used to illustrate the ability of the GNN to infer random walk models, are defined as follows: $C_{i,j} = \langle \delta_{\hat{m}(\Rt),i} \rangle_{\Rt\text{ of type j}}$, where $\hat{m}$ is the index of the inferred model, {i.e.} the one which has been assigned the highest probability, and $\delta$ is the Kronecker symbol.
That is, $C_{i,j}$ is the probability that a trajectory generated by the model class $j$ (column) is classified as belonging to model class $i$ (row).
Defined this way, the diagonal elements $C_{i,i}$ are the per-class recall.

\subsection{More complex Graph Neural Networks}
\label{sec:edge_features}

\begin{figure}
 \centering
  \includegraphics[width=0.45\linewidth]{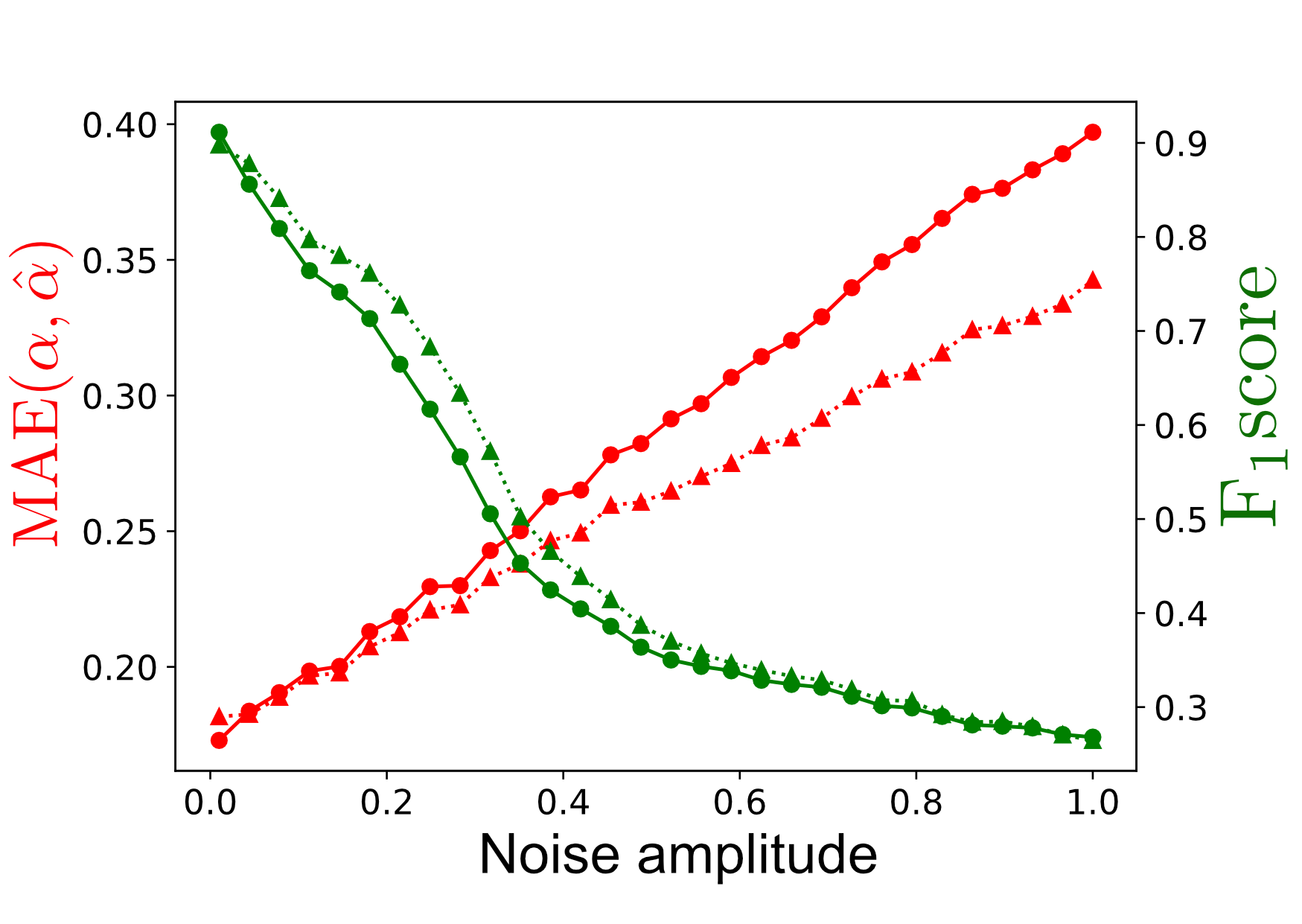}
  \caption{Performance  of GNNs trained with (plain lines, circles) or without (dashed lines, triangles) features attached to edges as a function of the noise amplitude in the trajectories. MAE is shown in red, $F_1$ score in green.}
  \label{fig:edge_features}
\end{figure}

In this paper, we focused on an approach where the GNN learns to build relevant features of random walks. We focused on a setting where features were only assigned to nodes, but the general framework of graph neural networks allows a more complex structure, by attaching features to edges, and by having two multi-layer perceptrons involved in the convolution operations : one before and one after the aggregation.
In this more general setting, the neural message passing equations ~\cite{Hamilton2020} read :
\begin{equation}
\vec{x}^{(k)}_{i} \leftarrow \gamma_k\pare{\vec{x}^{(k-1)}_{i}, \mathcal{X}_{j\in \mathcal{N}\pare{i}}\phi_k\pare{\vec{x}^{(k-1)}_{i}, \vec{x}^{(k-1)}_{j}, \vec{e}^{(k-1)}_{j,i}}  }	\enspace
\end{equation}
We show in Figure~\ref{fig:edge_features} that edge features allow better performance on noise-free trajectories but is less robust to localisation noise.

\subsection{Random versus structured connection patterns}
\label{random_vs_uniform}
\begin{figure}
 \centering
 \includegraphics[width=.8\linewidth]{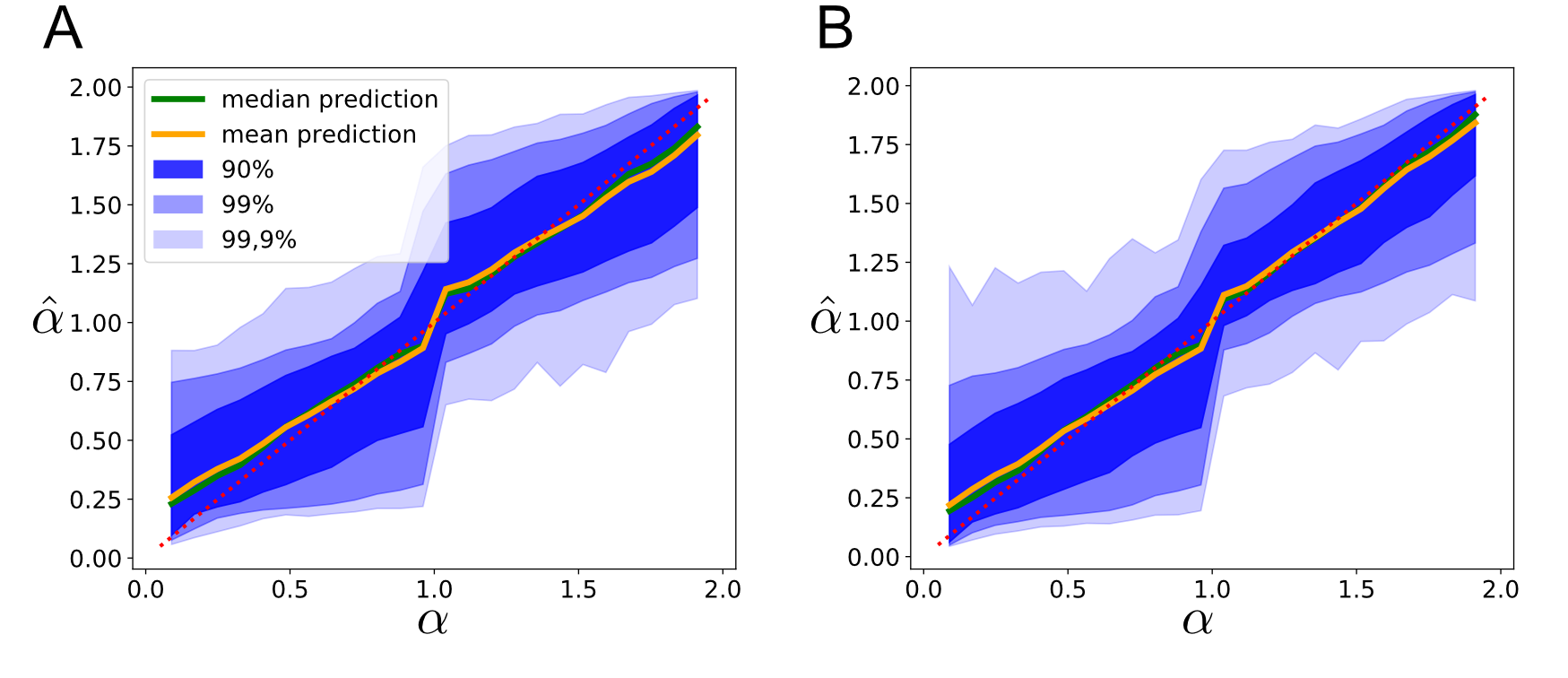}
 \caption{Regression performance of GNNs applied to different graph structures. 
 A) Random regular graph structure.
 B) Causal hierarchical graph structure.
  Shaded regions represent probability intervals of the estimators.}
 \label{fig:UnifGeom}
\end{figure}

As illustrated in Figure \ref{fig:UnifGeom}, connecting nodes according to a regular geometric pattern yields better performance than linking them at random.

In the geometric causal wiring scheme, node $i$ receives edges from nodes $i-\lfloor \beta_1 \rfloor, i-\lfloor \beta_2 \rfloor, \ldots, \min(0,i - \lfloor \beta_k \rfloor)$, where $k$ is the maximal \emph{in-}degree, independent of trajectory length and which enforces sparsity, and where $\beta_1, \beta_2, \ldots, \beta_k$ is a geometric progression, parametrized such that the last node receives a connection from the first node, and $\lfloor \beta \rfloor$ denotes the integer part of $\beta$. 
We ensure that no edge is doubled. 
We used a value of $k = 20$ throughout this paper.
Hence, nodes close to the start of the trajectory receive less connections than those located at the end. 

\renewcommand*{\bibfont}{\footnotesize}
\setlength{\bibitemsep}{0pt}
\printbibliography

\iftoggle{NAT}{}{

\appendix

\renewcommand{\thesection}{\Alph{section}}
\renewcommand{\thesubsection}{\Alph{section}\arabic{subsection}}
\renewcommand\thefigure{{\thesection}.\arabic{figure}}
\renewcommand{\theequation}{\Alph{section}.\arabic{equation}}
\renewcommand\thetable{\thesection.\arabic{table}}
\counterwithin{figure}{section}
\counterwithin{table}{section}
\counterwithin{equation}{section}

\maketitle

 \section{Appendix}

\begin{refsection}
\begingroup

\endgroup

\normalem
\renewcommand*{\bibfont}{\footnotesize}
\setlength{\bibitemsep}{0pt}
\printbibliography[heading=subbibliography]
\end{refsection}
}

\end{document}